\documentclass[10pt,twocolumn,letterpaper]{article}

\usepackage{cvpr}
\usepackage{times}
\usepackage{epsfig}
\usepackage{graphicx}
\usepackage{amsmath}
\usepackage{amssymb}
\usepackage{physics}
\usepackage{float}
\usepackage{booktabs}       % professional-quality tables

\usepackage{dblfloatfix}
\usepackage{transparent}

\usepackage{algorithm}
\usepackage{listings}

\newcommand*{\G}{\mathcal{G}}
\newcommand*{\V}{\mathcal{V}}
\newcommand*{\E}{\mathcal{E}}

\newcommand*{\I}{\mathcal{I}}
\newcommand*{\loss}{\mathcal{L}}

\newcommand*{\Z}{\mathcal{Z}}

\newcommand{\alg}{\texttt{edge-popup} }

\usepackage[pagebackref=true,breaklinks=true,letterpaper=true,colorlinks,bookmarks=false]{hyperref}

\cvprfinalcopy 

\def\cvprPaperID{6397} % *** Enter the CVPR Paper ID here

\cvprfinalcopy
\begin{document}

%%%%%%%%% TITLE
\title{What's Hidden in a Randomly Weighted Neural Network?}

\author{~~~~~~~~~~~ Vivek Ramanujan \thanks{equal contribution} \ \footnotemark[2]\and
Mitchell Wortsman \footnotemark[1] \ \footnotemark[3]\and
Aniruddha Kembhavi \thanks{Allen Institute for Artificial Intelligence} \ \thanks{University of Washington}~~~~~~~~~~~
\and
Ali Farhadi \ \footnotemark[3]
\and 
Mohammad Rastegari \ \footnotemark[3]
}

\maketitle

%%%%%%%%% ABSTRACT

\begin{abstract}
Training a neural network is synonymous with learning the values of the weights. In contrast, we demonstrate that randomly weighted neural networks contain subnetworks which achieve impressive performance without ever modifying the weight values. Hidden in a randomly weighted Wide ResNet-50 \cite{wideresnet} we find a subnetwork (with random weights) that is smaller than, but matches the performance of a ResNet-34 \cite{resnet} trained on ImageNet \cite{imagenet}. Not only do these ``untrained subnetworks" exist, but we provide an algorithm to effectively find them. We empirically show that as randomly weighted neural networks with fixed weights grow wider and deeper, an ``untrained subnetwork" approaches a network with learned weights in accuracy. Our code and pretrained models are available at: \href{https://github.com/allenai/hidden-networks}{https://github.com/allenai/hidden-networks}.
\end{abstract}
%%%%%%%%% INTRO
\section{Introduction}
\begin{figure}[t]
\begin{center}
\includegraphics[width=\columnwidth]{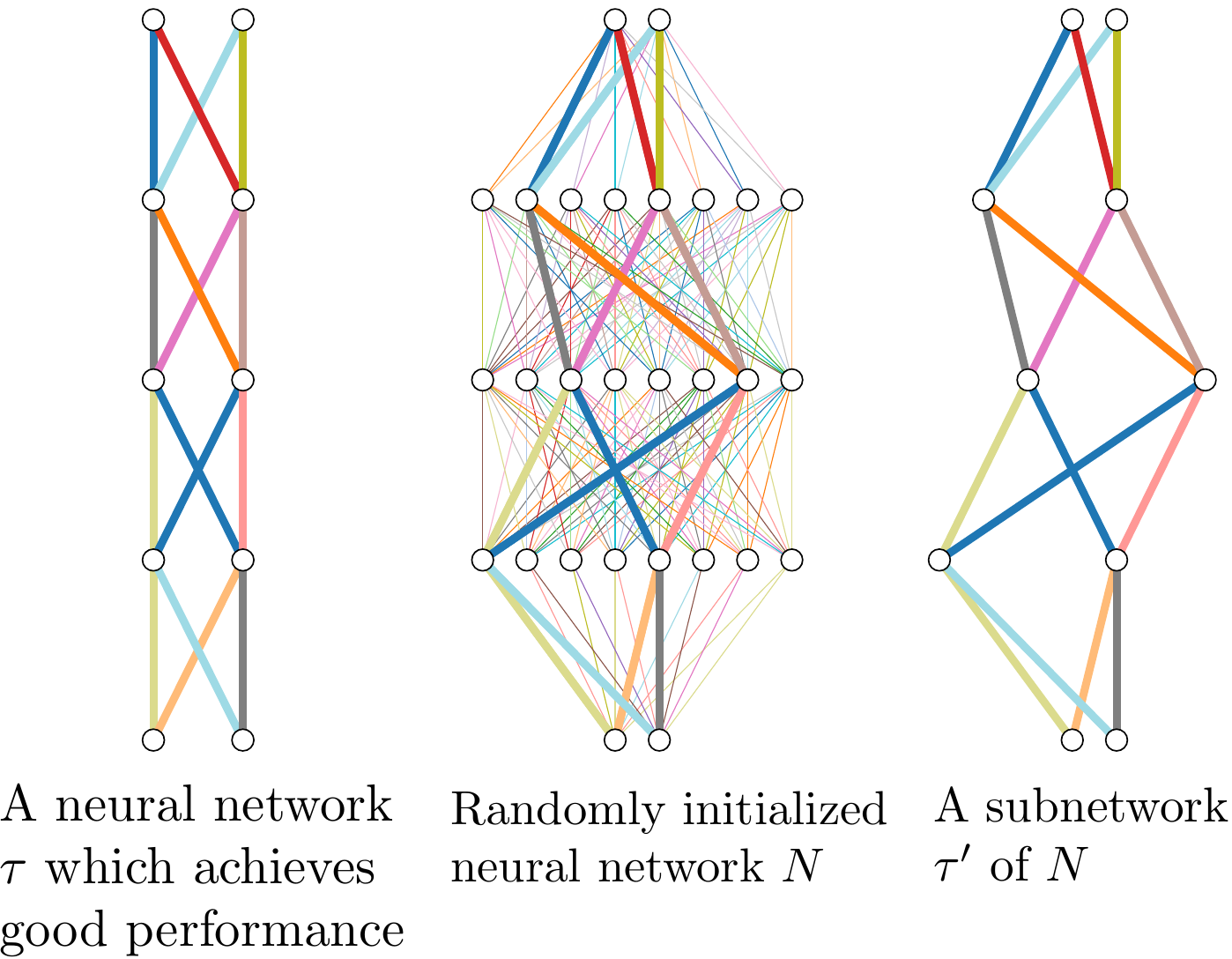}
\end{center}
  \caption{If a neural network with random weights (center) is sufficiently overparameterized, it will contain a subnetwork (right) that perform as well as a trained neural network (left) with the same number of parameters.}

\label{fig:teaser}
\end{figure}

What lies hidden in an overparameterized neural network with random weights? If the distribution is properly scaled, then it contains a subnetwork which performs well without ever modifying the values of the weights (as illustrated by Figure \ref{fig:teaser}).

The number of subnetworks is combinatorial in the size of the network, and modern neural networks contain millions or even billions of parameters \cite{2019t5}. We should expect that even a randomly weighted neural network contains a subnetwork that performs well on a given task. In this work, we provide an algorithm to find these subnetworks.

Finding subnetworks contrasts with the prevailing paradigm for neural network training -- learning the values of the weights by stochastic gradient descent. Traditionally, the network structure is either fixed during training (\eg ResNet \cite{resnet} or MobileNet \cite{mobilenet}), or optimized in conjunction with the weight values (\eg Neural Architecture Search (NAS)). We instead optimize to find a good subnetwork within a fixed, randomly weighted network. We do not ever tune the value of any weights in the network, not even the batch norm \cite{batchnorm} parameters or first or last layer.

In \cite{lth}, Frankle and Carbin articulate \textit{The Lottery Ticket Hypothesis}: neural networks contain sparse subnetworks that can be effectively trained from scratch when reset to their initialization.
We offer a complimentary conjecture: within a sufficiently overparameterized neural network with random weights (\eg at initialization), there exists a subnetwork that achieves competitive accuracy. 
Specifically, the test accuracy of the subnetwork is able to match the accuracy of a trained network with the same number of parameters. 

This work is catalyzed by the recent advances of Zhou \etal \cite{supermask}. By sampling subnetworks in the forward pass, they first demonstrate that subnetworks of randomly weighted neural networks can achieve impressive accuracy. However, we hypothesize that stochasticity may limit their performance. As the number of parameters in the network grows, they are likely to have a high variability in their sampled networks.

To this end we propose the \alg algorithm for finding effective subnetworks within randomly weighted neural networks. We show a significant boost in performance and scale to ImageNet. For each fixed random weight in the network, we consider a positive real-valued score. To choose a subnetwork we take the weights with the top-$k\%$ highest scores. With a gradient estimator we optimize the scores via SGD. We are therefore able to find a good neural network without ever changing the values of the weights. We empirically demonstrate the efficacy of our algorithm and show that (under certain technical assumptions) the loss decreases on the mini-batch with each modification of the subnetwork.

We experiment on small and large scale datasets for image recognition, namely CIFAR-10 \cite{cifar} and Imagenet \cite{imagenet}. On CIFAR-10 we empirically demonstrate that as networks grow wider and deeper, untrained subnetworks perform just as well as the dense network with learned weights. On ImageNet, we find a subnetwork of a randomly weighted Wide ResNet50 which is smaller than, but matches the performance of a trained ResNet-34. Moreover, a randomly weighted ResNet-101 \cite{resnet} with fixed weights contains a subnetwork that is much smaller, but surpasses the performance of VGG-16 \cite{vgg}. In short, we validate the unreasonable effectiveness of randomly weighted neural networks for image recognition.

\begin{figure*}[t!]
    \centering
    \includegraphics[width=\textwidth]{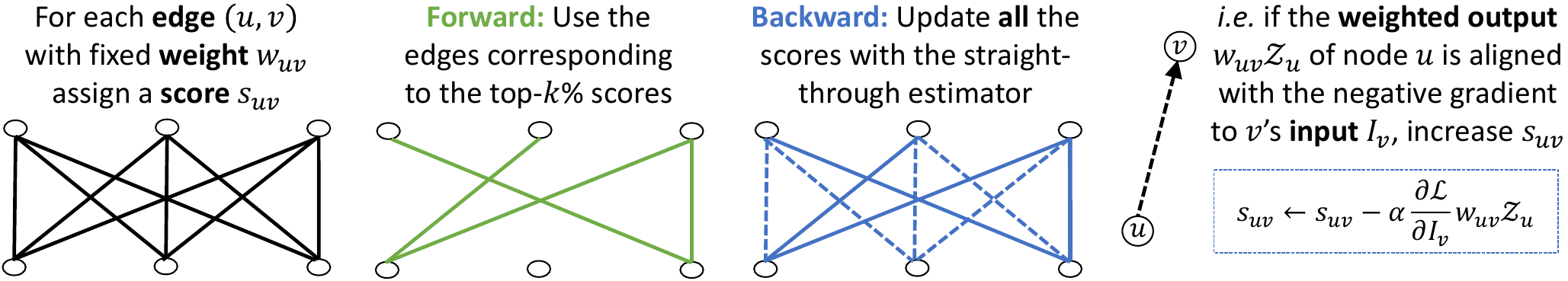}
    \caption{In the \alg Algorithm, we associate a score with each edge. On the forward pass we choose the top edges by score. On the backward pass we update the scores of all the edges with the straight-through estimator, allowing helpful edges that are ``dead" to re-enter the subnetwork. We never update the value of any weight in the network, only the score associated with each weight.}
    \label{fig:alg}
\end{figure*}

%%%%%%%%% RELATED WORK
\section{Related Work}

\noindent\textbf{Lottery Tickets and Supermasks}

In \cite{lth}, Frankle and Carbin offer an intriguing hypothesis: neural networks contain sparse subnetworks that can be effectively trained from scratch when reset to their initialization. These so-called winning tickets have won the ``initialization lottery". Frankle and Carbin find winning tickets by iteratively shrinking the size of the network, masking out weights which have the lowest magnitude at the end of each training run.

Follow up work by Zhou \etal \cite{supermask} demonstrates that winning tickets achieve better than random performance without training. Motivated by this result they propose an algorithm to identify a ``supermask" -- a subnetwork of a randomly initialized neural network that achieves high accuracy without training. On CIFAR-10, they are able to find subnetworks of randomly initialized neural networks that achieve 65.4\% accuracy.

The algorithm presented by Zhou \etal is as follows: for each weight $w$ in the network they learn an associated probability $p$. On the forward pass they include weight $w$ with probability $p$ and otherwise zero it out. Equivalently, they use weight $\tilde w = w X$ where $X$ is a Bernoulli$(p)$ random variable ($X$ is $1$ with probability $p$ and $0$ otherwise). The probabilities $p$ are the output of a sigmoid, and are learned using stochastic gradient descent. The terminology supermask" arises as finding a subnetwork is equivalent to learning a binary mask for the weights.

Our work builds upon Zhou \etal, though we recognize that the stochasticity of their algorithm may limit performance. In section \ref{sec:int} we provide more intuition for this claim. We show a significant boost in performance with an algorithm that does not sample supermasks on the forward pass. For the first time we are able to match the performance of a dense network with a supermask.

\noindent\textbf{Neural Architecture Search (NAS)}

The advent of modern neural networks has shifted the focus from feature engineering to feature learning. However, researchers may now find themselves manually engineering the architecture of the network. Methods of Neural Architecture Search (NAS) \cite{nas, cai2018proxylessnas, liu2018progressive, mnasnet} instead provide a mechanism for learning the architecture of neural network jointly with the weights. Models powered by NAS have recently obtained state of the art classification performance on ImageNet \cite{efficientnet}.

As highlighted by Xie \etal \cite{xie2019exploring}, the connectivity patterns in methods of NAS remain largely constrained. Surprisingly, Xie \etal establish that randomly wired neural networks can achieve competitive performance. Accordingly, Wortsman \etal \cite{dnw} propose a method of Discovering Neural Wirings (DNW) -- where the weights and structure are jointly optimized free from the typical constraints of NAS. We highlight DNW as we use a similar method of analysis and gradient estimator to optimize our supermasks. In DNW, however, the subnetwork is chosen by taking the weights with the highest magnitude. There is therefore no way to learn supermasks with DNW as the weights and connectivity are inextricably linked -- there is no way to separate the weights and the structure.

\noindent\textbf{Weight Agnostic Neural Networks}

In Weight Agnostic Neural Networks (WANNs) \cite{wann2019}, Gaier and Ha question if an architecture alone may encode the solution to a problem. They present a mechanism for building neural networks that achieve high performance when each weight in the network has the same shared value. Importantly, the performance of the network is agnostic to the value itself. They are able to obtain $\sim 92\%$ accuracy on MNIST \cite{mnist}.

We are quite inspired by WANNs, though we would like to highlight some important distinctions. Instead of each weight having the same value, we explore the setting where each weight has a random value. In Section A.2.2 of their appendix, Gaier and Ha mention that they were not successful in this setting. However, we find a good subnetwork for a given random initialization -- the supermasks we find are not agnostic to the weights. Finally, Gaier and Ha construct their network architectures, while we look for supermasks within standard architectures.

\noindent\textbf{Linear Classifiers and Pruning at Initialization}

Linear classifiers on top of randomly weighted neural networks are often used as baselines in unsupervised learning \cite{jigsaw, elm}. This work is different in motivation, we search for untrained subnetworks which achieve high performance without changing any weight values. This also differs from methods which prune at initialization and modify the weights of the discovered subnetwork \cite{lee2018snip, lee2019signal} or methods which modify a subset of the weights \cite{rosenfeld2019intriguing}.

\section{Method}

In this section we present our optimization method for finding effective subnetworks within randomly weighted neural networks. We begin by building intuition in an unusual setting -- the infinite width limit. Next we motivate and present our algorithm for finding effective subnetworks.

\subsection{Intuition} \label{sec:int}

\noindent \textbf{The Existence of Good Subnetworks} 

Modern neural networks have a staggering number of possible subnetworks. Consequently, even at initialization, a neural network should contain a subnetwork which performs well. 

To build intuition we will consider an extreme case -- a neural network $N$ in the infinite width limit (for a convolutional neural networks, the width of the network is the number of channels). As in Figure \ref{fig:teaser}, let $\tau$ be a network with the same structure of $N$ that achieves good accuracy. If the weights of $N$ are initialized using any standard scaling of a normal distribution, \eg xavier \cite{glorot} or kaiming \cite{kaiming-init}, then we may show there exists a subnetwork of $N$ that achieves the same performance as $\tau$ without training. Let $q$ be the probability that a given subnetwork of $N$ has weights that are close enough to $\tau$ to obtain the same accuracy. This probability $q$ is extremely small, but it is still nonzero. Therefore, the probability that no subnetwork of $N$ is close enough to $\tau$ is effectively $(1-q)^S$ where $S$ is the number of subnetworks. $S$ grows very quickly with the width of the network, and this probability becomes arbitrarily small.

\noindent \textbf{How Should We Find A Good Subnetwork}

Even if there are good subnetworks in randomly weighted neural networks, how should we find them?

Zhou \etal learn an associated probability $p$ with each weight $w$ in the network. On the forward pass they include weight $w$ with probability $p$ (where $p$ is the output of a sigmoid) and otherwise zero it out. The infinite width limit provides intuition for a possible shortcoming of the algorithm presented by Zhou \etal \cite{supermask}. Even if the parameters $p$ are fixed, the algorithm will likely never observe the same subnetwork twice. As such, the gradient estimate becomes more unstable, and this in turn may make training difficult.

Our algorithm for finding a good subnetwork is illustrated by Figure \ref{fig:alg}. With each weight $w$ in the neural network we learn a positive, real valued \textit{popup} score $s$. The subnetwork is then chosen by selecting the weights in each layer corresponding to the top-$k\%$ highest scores. For simplicity we use the same value of $k$ for all layers. 

How should we update the score $s_{uv}$? Consider a single edge in a fully connected layer which connects neuron $u$ to neuron $v$. Let $w_{uv}$ be the weight of this edge, and $s_{uv}$ the associated score. If this score is initially low then $w_{uv}$ is not selected in the forward pass. But we would still like a way to update its score to allow it to pop back up. Informally, with backprop \cite{backprop} we compute how the loss ``wants" node $v$'s input to change (\ie the negative gradient). We then examine the weighted output of node $u$. If this weighted output is aligned with the negative gradient, then node $u$ can take node $v$'s output where the loss ``wants" it to go. Accordingly, we should increase the score. If this alignment happens consistently, then the score will continue to increase and the edge will re-enter the chosen subnetwork (\ie popup).

More formally, if $w_{uv}\Z_{u}$ denotes the weighted output of neuron $u$, and $\I_v$ denotes the input of neuron $v$, then we update $s_{uv}$ as 
\begin{align}
    s_{uv} \gets s_{uv} - \alpha \frac{\partial \loss}{\partial \I_v} \Z_u w_{uv}.
\end{align}

This argument and the analysis that follows is motivated and guided by the work of \cite{dnw}. In their work, however, they do not consider a score and are instead directly updating the weights. In the forward pass they use the top $k\%$ of edges by magnitude, and therefore there is no way of learning a subnetwork without learning the weights. Their goal is to train sparse neural networks, while we aim to showcase the efficacy of randomly weighted neural networks.

\subsection{The \alg Algorithm and Analysis} \label{sec:alg}

We now formally detail the \alg algorithm. 

For clarity, we first describe our algorithm for a fully connected neural network. In Section \ref{sec:convolutional-case} we provide the straightforward extension to convolutions along with code in PyTorch \cite{pytorch}.

A fully connected neural network consists of layers $1,...,L$ where layer $\ell$ has $n_\ell$ nodes $\V^{(\ell)} = \{v^{(\ell)}_1,..., v^{(\ell)}_{n_\ell}\}$. We let $\I_v$ denote the input to node $v$ and let $\Z_v$ denote the output, where $\Z_v = \sigma(\I_v)$ for some non-linear activation function $\sigma$ (\eg ReLU \cite{alexnet}). The input to neuron $v$ in layer $\ell$ is a weighted sum of all neurons in the preceding layer. Accordingly, we write $\I_v$ as
\begin{align} \label{eq:linear}
    \I_{v} = \sum_{u \in \V^{(\ell - 1)}} w_{uv} \Z_u
\end{align}
where $w_{uv}$ are the network parameters for layer $\ell$. The output of the network is taken from the final layer while the input data is given to the very first layer. Before training, the weights $w_{uv}$ for layer $\ell$ are initialized by independently sampling from distribution $\mathcal{D}_\ell$. For example, if we are using kaiming normal initialization \cite{kaiming-init} with ReLU activations, then $\mathcal{D}_\ell = \mathcal{N}\left(0, \sqrt{2/n_{\ell -1}} \right)$ where $\mathcal{N}$ denotes the normal distribution.

Normally, the weights $w_{uv}$ are optimized via stochastic gradient descent. In our \alg algorithm, we instead keep the weights at their random initialization, and optimize to find a subnetwork $\G = (\V, \E)$. We then compute the input of node $v$ in layer $\ell$ as 
\begin{align} \label{eq:our-update}
    \I_{v} = \sum_{(u,v) \in \E} w_{uv} \Z_u
\end{align}
where $\G$ is a subgraph of the original fully connected network\footnote{The original network has edges $\E_\textrm{fc} = \bigcup_{\ell = 1}^{L-1} \left( \V_{\ell} \times \V_{\ell + 1}\right)$ where $\times$ denotes the cross-product.}.  As mentioned above, for each weight $w_{uv}$ in the original network we learn a \textit{popup} score $s_{uv}$. We choose the subnetwork $\G$ by selecting the weights in each layer which have the top-$k\%$ highest scores. Equation \ref{eq:our-update} may therefore be written equivalently as 
\begin{align} \label{eq:our-update2}
    \I_{v} = \sum_{u \in \V^{(\ell - 1)}} w_{uv} \Z_u h(s_{uv})
\end{align}
where $h(s_{uv}) = 1$ if $s_{uv}$ is among the top $k\%$ highest scores in layer $\ell$ and $h(s_{uv}) = 0$ otherwise. Since the gradient of $h$ is 0 everywhere it is not possible to directly compute the gradient of the loss with respect to $s_{uv}$. We instead use the straight-through gradient estimator \cite{ste}, in which $h$ is treated as the identity in the backwards pass -- the gradient goes ``straight-through" $h$. Consequently, we approximate the gradient to $s_{uv}$ as
\begin{align}
    \hat g_{s_{uv}} = \pdv{\loss}{\I_v}\pdv{\I_v}{s_{uv}} = \pdv{\loss}{\I_v} w_{uv}\Z_u
\end{align}
where $\loss$ is the loss we are trying to minimize. The scores $s_{uv}$ are then updated via stochastic gradient descent with learning rate $\alpha$. If we ignore momentum and weight decay \cite{wd} then we update $s_{uv}$ as
\begin{align} \label{eq:update-rule}
    \tilde s_{uv} = s_{uv} - \alpha \pdv{\loss}{\I_v} w_{uv}\Z_u
\end{align}
where $\tilde s_{uv}$ denotes the score after the gradient step\footnote{To ensure that the scores are positive we take the absolute value.}.

As the scores change certain edges in the subnetwork will be replaced with others. Motivated by the analysis of \cite{dnw} we show that when swapping does occur, the loss decreases for the mini-batch. 

\noindent \textbf{Theorem 1:} When edge $(i,\rho)$ replaces $(j,\rho)$ and the rest of the subnetwork remains fixed then the loss decreases for the mini-batch (provided the loss is sufficiently smooth).

\noindent \textit{Proof.} Let $\tilde s_{uv}$ denote the score of weight $w_{uv}$ after the gradient update. If edge $(i,\rho)$ replaces $(j,\rho)$ then our algorithm dictates that $s_{i\rho} < s_{j\rho}$ but $\tilde s_{i\rho} > \tilde s_{j\rho}$. Accordingly,
\begin{align}
    &\tilde s_{i\rho} - s_{i\rho} > \tilde s_{j\rho} - s_{j\rho}
\end{align}
which implies that
\begin{align} \label{eq:ineq}
- \alpha \pdv{\loss}{\I_\rho} w_{i\rho}\Z_i > - \alpha \pdv{\loss}{\I_\rho} w_{j\rho}\Z_j
\end{align}
by the update rule given in Equation \ref{eq:update-rule}.
Let $\tilde \I_\rho$ denote the input to node $k$ after the swap is made and $\I_\rho$ denote the original input. Note that $\tilde \I_\rho - \I_\rho = w_{i\rho}Z_i - w_{j\rho}Z_j$ by Equation \ref{eq:our-update}. We now wish to show that $\loss(\tilde \I_\rho) < \loss\left(\I_\rho\right)$. 

If the loss is smooth and $ \tilde \I_\rho$ is close to $\I_\rho$ and ignore second-order terms in a Taylor expansion:
\begin{align}
    \loss\left(\tilde \I_\rho\right) &= \loss\left(\I_\rho + \left(\tilde \I_\rho -  \I_\rho\right)\right) \\
    &\approx \loss\left(\I_\rho \right) + \frac{\partial \loss}{\partial \I_\rho} \left( \tilde \I_\rho - \I_\rho\right)\\
    &= \loss\left(\I_\rho \right) + \frac{\partial \loss}{\partial \I_\rho} (w_{i\rho}\Z_i - w_{j\rho}\Z_j)
\end{align}
and from equation \ref{eq:ineq} we have that $\frac{\partial \loss}{\partial \I_\rho} (w_{i\rho}\Z_i - w_{j\rho}\Z_j) < 0$ and so $\loss(\tilde \I_\rho) < \loss\left(\I_\rho\right)$ as needed. We examine a more general case of Theorem 1 in Section \ref{sec:general-case} of the appendix.

\begin{figure*}[t!]
    \centering
    \includegraphics[width=\textwidth]{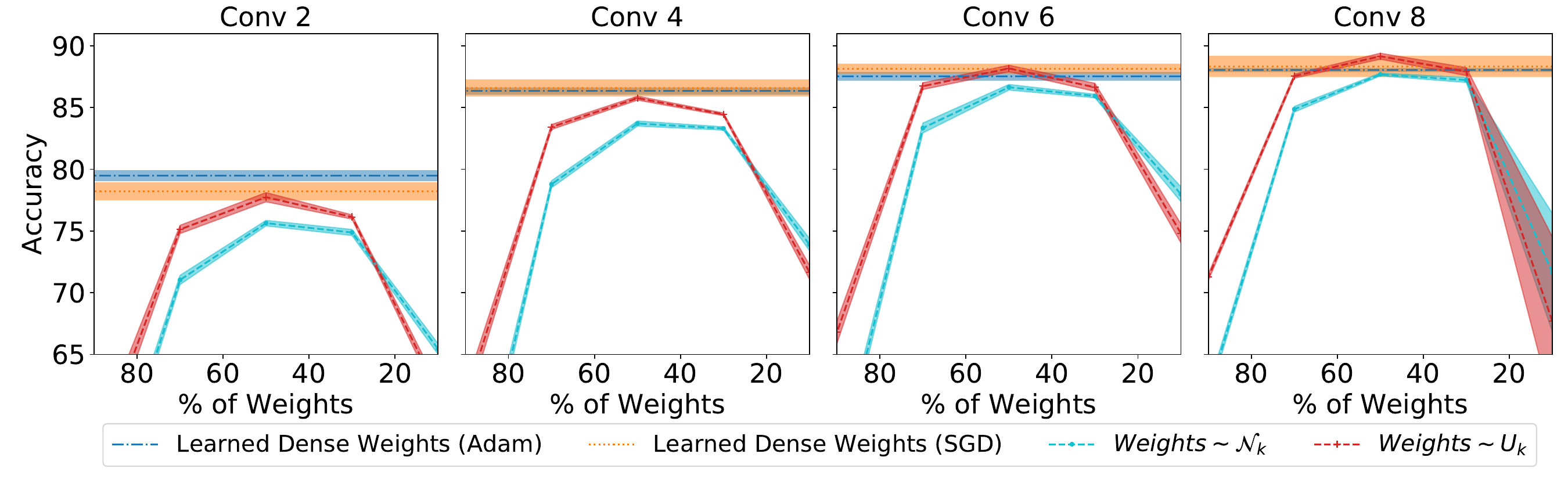}
    \caption{\textbf{Going Deeper:} Experimenting with shallow to deep neural networks on CIFAR-10 \cite{cifar}. As the network becomes deeper, we are able to find subnetworks at initialization that perform as well as the dense original network when trained. The baselines are drawn as a horizontal line as we are not varying the \% of weights. When we write \textit{Weights} $\sim \mathcal{D}$ we mean that the weights are randomly drawn from distribution $\mathcal{D}$ and are never tuned. Instead we find subnetworks with size (\% of Weights)$/100$ * (Total \# of Weights).}
    \label{fig:conv2468}
\end{figure*}

\section{Experiments}\label{sec:exp}

We demonstrate the efficacy of randomly weighted neural networks for image recognition on standard benchmark datasets CIFAR-10 \cite{cifar} and ImageNet \cite{imagenet}. This section is organized as follows: in Section \ref{sec:setup} we discuss the experimental setup and hyperparameters. We perform a series of ablations at small scale: we examine the effect of $k$, the $\%$ of Weights which remain in the subnetwork, and the effect of width. In Section \ref{sec:baseline} we compare against the algorithm of Zhou \etal, followed by Section \ref{sec:init} in which we study the effect of the distribution used to sample the weights. We conclude with Section \ref{sec:imagenet}, where we optimize to find subnetworks of randomly weighted neural networks which achieve good performance on ImageNet.

\subsection{Experimental Setup} \label{sec:setup}

We use two different distributions for the weights in our network:
\begin{itemize}
    \item \textbf{Kaiming Normal} \cite{kaiming-init}, which we denote $\mathcal{N}_k$. Following the notation in section \ref{sec:alg}  the Kaiming Normal distribution is defined as $\mathcal{N}_k = \mathcal{N}\left(0, \sqrt{2/n_{\ell -1}} \right)$ where $\mathcal{N}$ denotes the normal distribution.
    \item \textbf{Signed Kaiming Constant} which we denote $U_k$. Here we set each weight to be a constant and randomly choose its sign to be $+$ or $-$. The constant we choose is the standard deviation of Kaiming Normal, and as a result the variance is the same. We use the notation $U_k$ as we are sampling uniformly from the set $\{-\sigma_k, \sigma_k\}$ where $\sigma_k$ is the standard deviation for Kaiming Normal (\ie $\sqrt{2/n_{\ell -1}}$).
\end{itemize}
In Section \ref{sec:init} we reflect on the importance of the random distribution and experiment with alternatives.

On CIFAR-10 \cite{cifar} we experiment with simple VGG-like architectures of varying depth. These architectures are also used by Frankle and Carbin \cite{lth} and Zhou \etal \cite{supermask} and are provided in Table \ref{tab:cifarmodels}.  On ImageNet we experiment with ResNet-50 and ResNet-101 \cite{resnet}, as well as their wide variants \cite{wideresnet}.
\begin{table}
\scriptsize
\centering
\begin{tabular}{c  c  c  c  c}\toprule
Model & Conv2 & Conv4 & Conv6 & Conv8 \\ \midrule
\shortstack{Conv \\ Layers} &
\shortstack{64, 64, pool}  &
\shortstack{64, 64, pool\\128, 128, pool}  &
\shortstack{64, 64, pool\\128, 128, pool\\256, 256, pool}  &
\shortstack{64, 64, pool\\128, 128, pool\\256, 256, pool\\512, 512, pool}  
\\ \midrule
\shortstack{FC} & 256, 256, 10 & 256, 256, 10 & 256, 256, 10 & 256, 256, 10 \\ 
\bottomrule
\\
\end{tabular}
\caption{For completeness we provide the architecture of the simple VGG-like \cite{vgg} architectures used for CIFAR-10 \cite{cifar}, which are identical to those used by Frankle and Carbin \cite{lth} and Zhou \etal \cite{supermask}. However, the slightly deeper Conv8 does not appear in the previous work. Each model first performs convolutions followed by the fully connected (FC) layers, and pool denotes max-pooling.}
\label{tab:cifarmodels}
\end{table} In every experiment (for all baselines, datasets, and our algorithm) we optimize for 100 epochs and report the last epoch accuracy on the validation set. When we optimize with Adam \cite{adam} we do not decay the learning rate. When we optimize with SGD we use cosine learning rate decay \cite{cosinelr}.  On CIFAR-10 \cite{cifar} we train our models with weight decay 1e-4, momentum 0.9,  batch size 128, and learning rate 0.1. We also often run both an Adam and SGD baseline where the weights are learned. The Adam baseline uses the same learning rate and batch size as in \cite{lth, supermask}\footnote{Batch size 60, learning rate 2e-4, 3e-4 and 3e-4 for Conv2, Conv4, and Conv6 respectively Conv8 is not tested in \cite{lth}, though we use find that learning rate 3e-4 still performs well.}. For the SGD baseline we find that training does not converge with learning rate 0.1, and so we use 0.01. As standard we also use weight decay 1e-4, momentum 0.9, and batch size 128.
\begin{figure*}[t!]
    \centering
    \includegraphics[width=\textwidth]{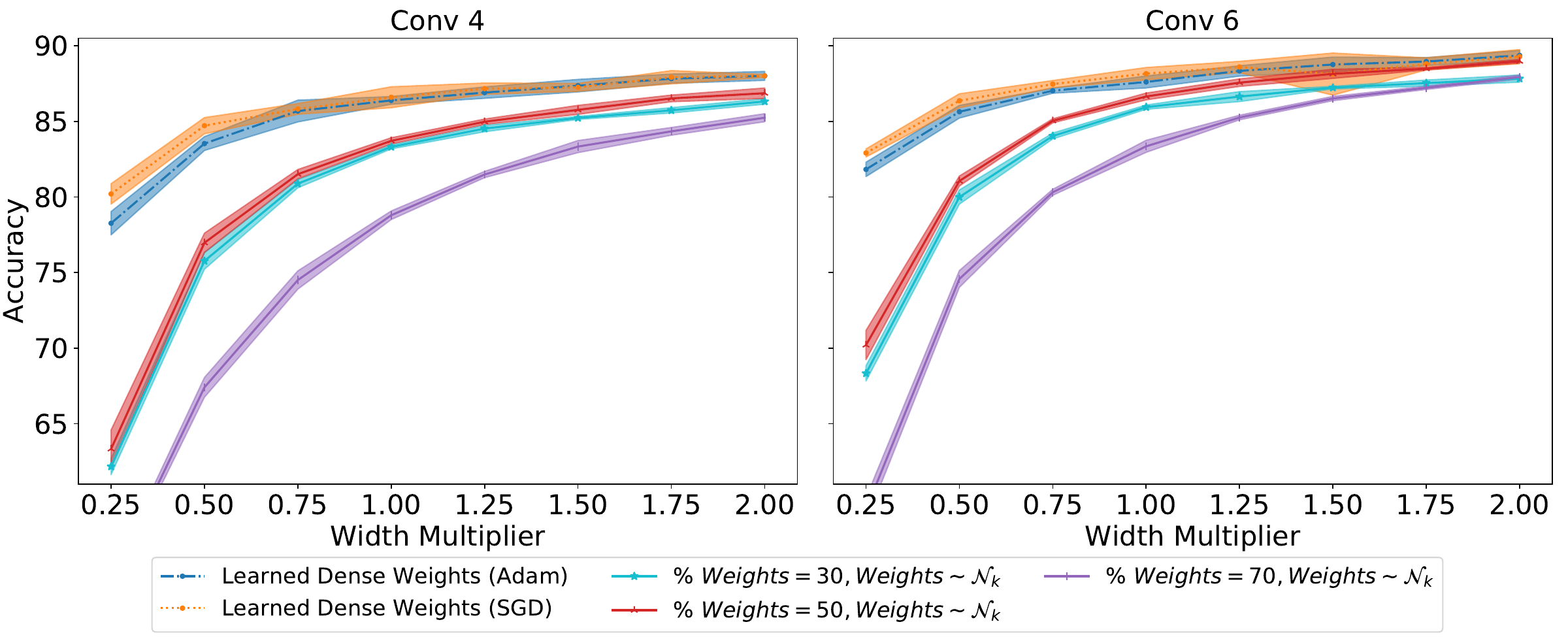}
    \caption{\textbf{Going Wider:} Varying the width (\ie number of channels) of Conv4 and Conv6 for CIFAR-10 \cite{cifar}. When Conv6 is wide enough, a subnetwork of the randomly weighted model (with $\% \textrm{\textit{Weights}} = 50$) performs just as well as the full model when it is trained.}
    \label{fig:wide}
\end{figure*}
For the ImageNet experiments we use the hyperparameters found on NVIDIA's public github example repository for training ResNet \cite{repo}. For simplicity, our \alg algorithm does not modify batch norm parameters, they are frozen at their default initialization in PyTorch (\ie bias 0, scale 1), and the scores are initialized Kaiming uniform \cite{kaiming-init}.

This discussion has encompassed the extent of the hyperparameter tuning for our models. We do, however, perform hyperparameter tuning for the Zhou \etal \cite{supermask} baseline and improve accuracy significantly. We include further discussion of this in Section \ref{sec:baseline}.

In all experiments on CIFAR-10 \cite{cifar} we use 5 different random seeds and plot the mean accuracy $\pm$ one standard deviation. Moreover, on all figures, \textit{Learned Dense Weights} denotes the standard training the full model (all weights remaining).

\subsection{Varying the \% of Weights}

Our algorithm has one associated parameter: the \% of weights which remain in the subnetwork, which we refer to as $k$. Figure \ref{fig:conv2468} illustrates how the accuracy of the subnetwork we find varies with $k$, a trend which we will now dissect. We consider $k \in [10, 30, 50, 70, 90]$ and plot the dense model when it is trained as a horizontal line (as it has 100\% of the weights). 

We recieve the worst accuracy when $k$ approaches $0$ or $100$. When $k$ approaches 0, we are not able to perform well as our subnetwork has very few weights. On the other hand, when $k$ approaches 100, our network outputs are random.

The best accuracy occurs when $k \in [30, 70]$, and we make a combinatorial argument for this trend. We are choosing $kn$ weights out of $n$, and there are $n \choose kn$ ways of doing so. The number of possible subnetworks is therefore maximized when $k \approx 0.5 $, and at this value our search space is at its largest.

\subsection{Varying the Width}

Our intuition from Section \ref{sec:int} suggests that as the network gets wider, a subnetwork of a randomly weighted model should approach the trained model in accuracy. How wide is wide enough? 
\begin{figure}[h]
    \centering
    \includegraphics[width=0.9\columnwidth]{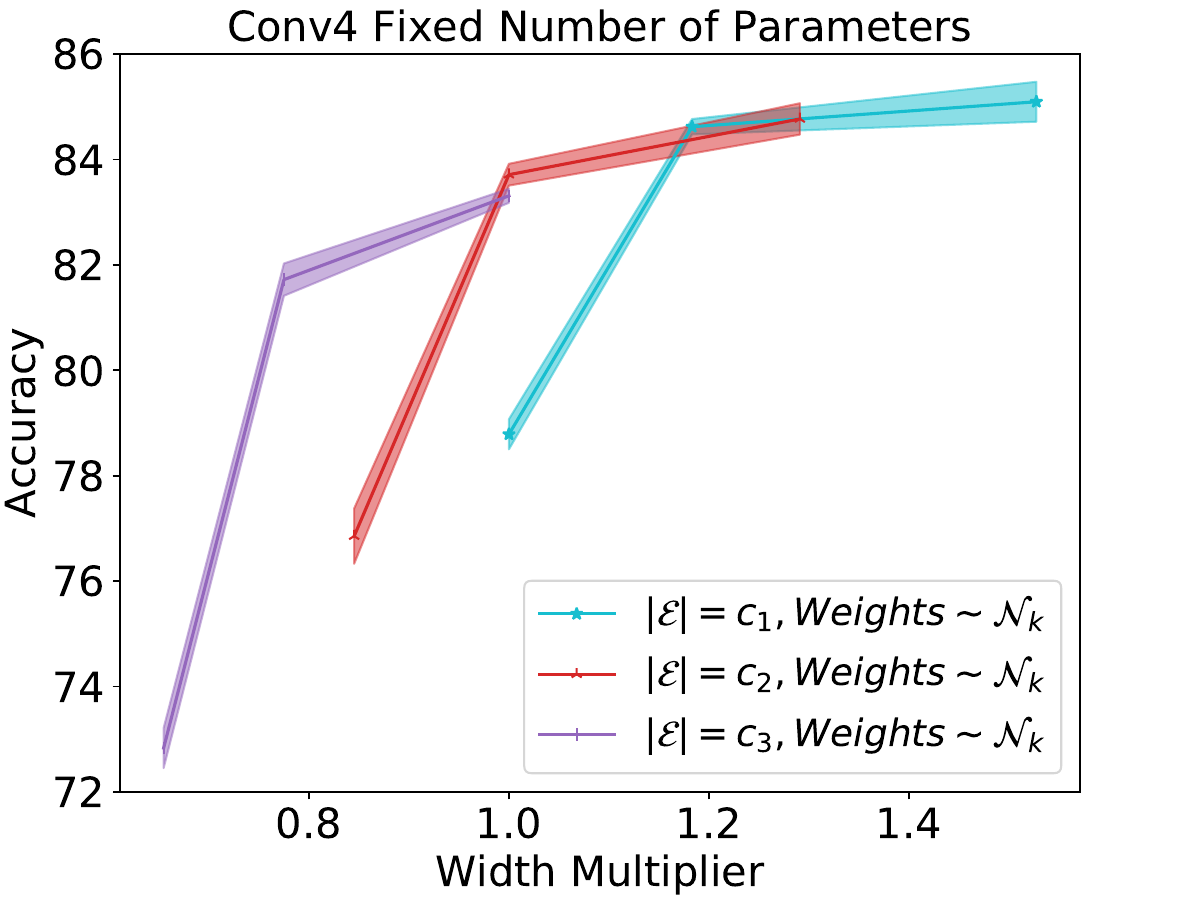}
    \caption{Varying the width of Conv4  on CIFAR-10 \cite{cifar} while modifying $k$ so that the \# of Parameters is fixed along each curve. $c_1,c_2, c_3$ are constants which coincide with \# of Parameters for $k=[30,50,70]$ for width multiplier 1.}
    \label{fig:eqparams}
\end{figure}

\begin{figure*}[t!]
    \centering
    \includegraphics[width=\textwidth]{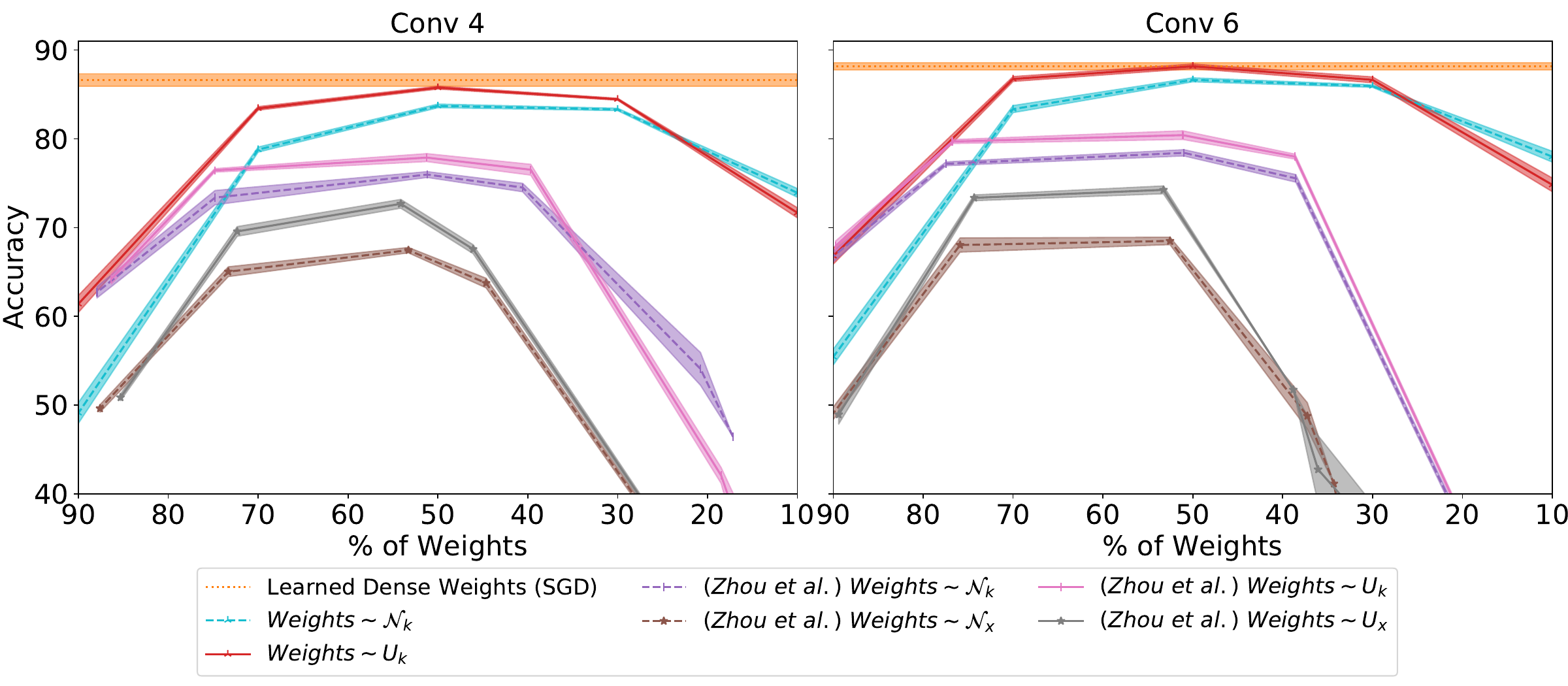}
    \caption{Comparing the performance of \alg with the algorithm presented by Zhou \etal \cite{supermask} on CIFAR-10 \cite{cifar}.}
    \label{fig:zhou}
\end{figure*}

In Figure \ref{fig:wide} we vary the width of Conv4 and Conv6.  The width of a linear layer is the number of ``neurons", and the width of a convolution layer is the number of channels. The width multiplier is the factor by which the width of all layers is scaled. A width multiplier of 1 corresponds to the models tested in Figure \ref{fig:conv2468}.

As the width multiplier increases, the gap shrinks between the accuracy a subnetwork found with \alg and the dense model when it is trained. Notably, when Conv6 is wide enough, a subnetwork of the randomly weighted model (with $\% \textrm{\textit{Weights}} = 50$) performs just as well as the dense model when it is trained. 

Moreover, this boost in performance is not solely from the subnetwork having more parameters. Even when the \# of parameters is fixed, increasing the width and therefore the search space leads to better performance. In Figure \ref{fig:eqparams} we fix the number of parameters and while modifying $k$ and the width multiplier. Specifically, we test $k \in [30, 50, 70]$ for subnetworks of constant size $c_1, c_2$ and $c_3$. On Figure \ref{fig:eqparams} we use $|\E|$ denote the size of the subnetwork.

\subsection{Comparing with Zhou \etal \cite{supermask}} \label{sec:baseline}

In Figure \ref{fig:zhou} we compare the performance of \alg with Zhou \etal. Their work considers distributions $\mathcal{N}_x$ and $U_x$, which are identical to those presented in Section \ref{sec:setup} but with xavier normal \cite{glorot} instead of kaiming normal \cite{kaiming-init} -- the factor of $\sqrt{2}$ is omitted from the standard deviation. By running their algorithm with $\mathcal{N}_k$ and $U_k$ we witness a significant improvement. However, even the $\mathcal{N}_x$ and $U_x$ results exceed those in the paper as we perform some hyperparameter tuning. As in our experiments on CIFAR-10, we use SGD with weight decay 1e-4, momentum 0.9, batch size 128, and a cosine scheduler \cite{cosinelr}. We double the learning rate until we see the performance become worse, and settle on 200\footnote{A very high learning rate is required as mentioned in their work.}.

\begin{figure}[h]
    \centering
    \includegraphics[width=\columnwidth]{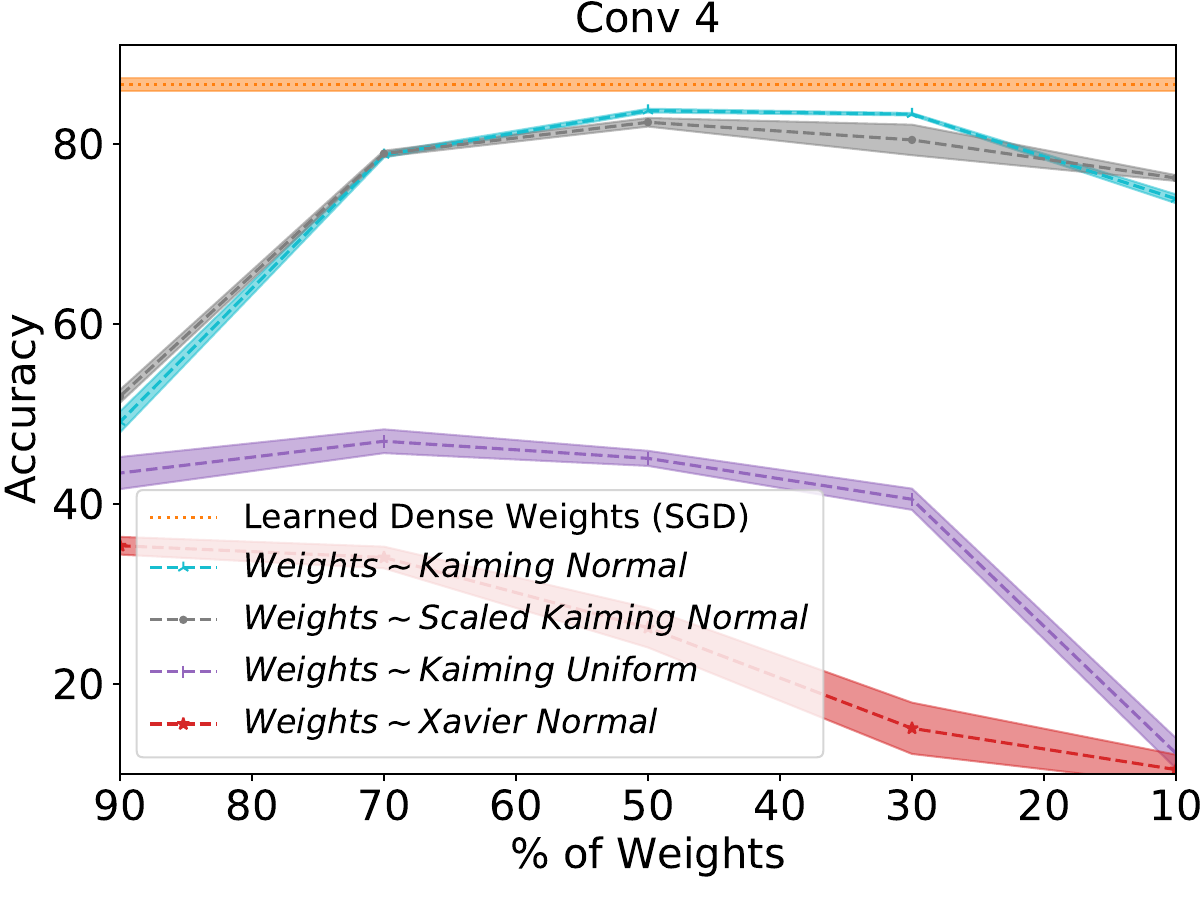}
    \caption{Testing different weight distributions on CIFAR-10 \cite{cifar}.}
    \label{fig:init}
\end{figure}

\subsection{Effect of The Distribution} \label{sec:init}

The distribution that the random weights are sampled from is very important. As illustrated by Figure \ref{fig:init}, the performance of our algorithm vastly decreases when we switch to using xavier normal \cite{glorot} or kaiming uniform \cite{kaiming-init}.

Following the derivation in \cite{kaiming-init}, the variance of the forward pass is not exactly 1 when we consider a subnetwork with only $k\%$ of the weights. To reconcile for this we could scale standard deviation by $\sqrt{1/k}$. This distribution is referred to as ``Scaled Kaiming Normal" on Figure \ref{fig:init}. We may also consider this scaling for the Signed Kaiming Constant distribution which is described in Section \ref{sec:setup}.

\begin{figure}[h]
\begin{center}
\includegraphics[width=0.9\columnwidth]{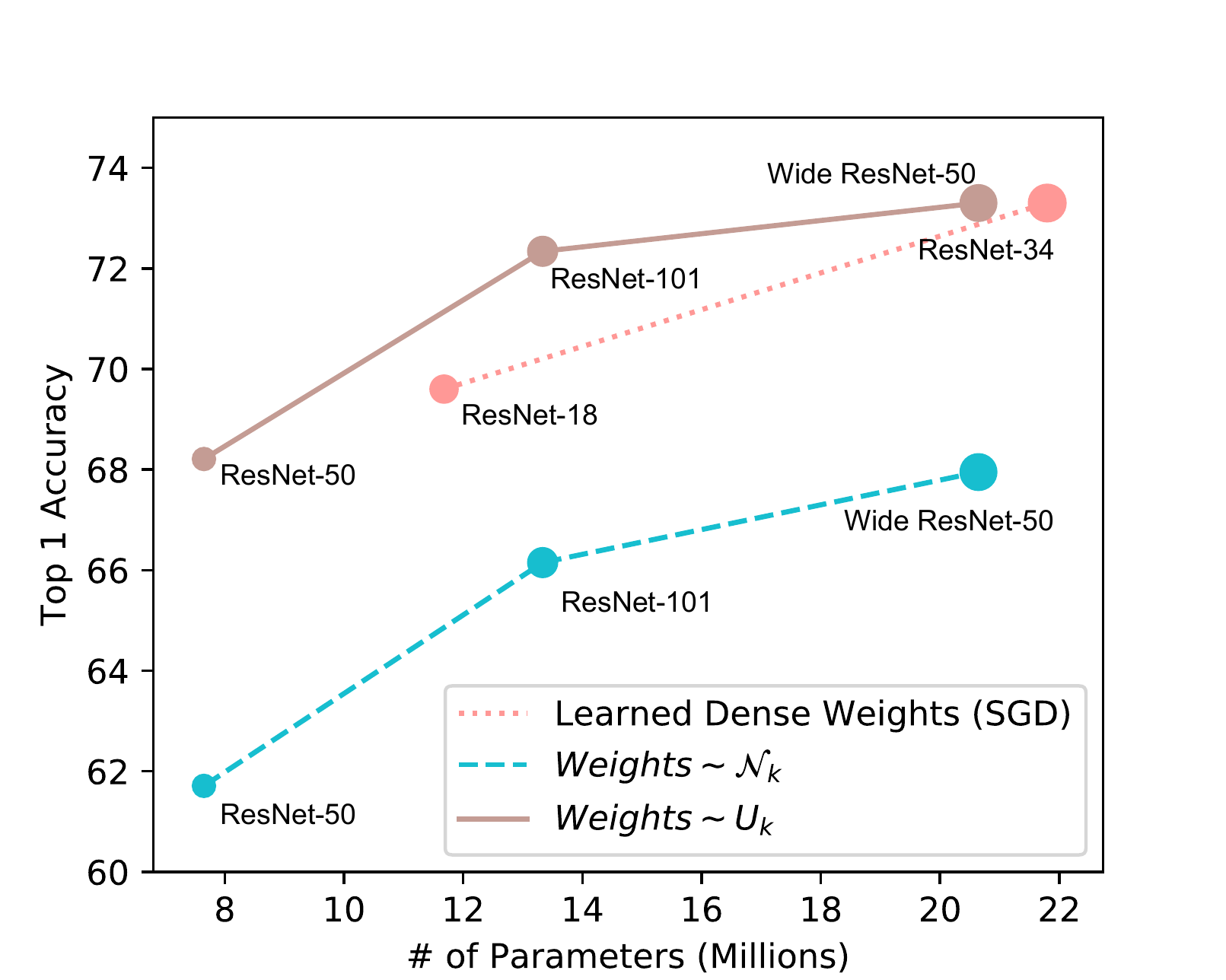}
  
\end{center}
  \caption{Testing our Algorithm on ImageNet \cite{imagenet}. We use a fixed $k = 30\%$, and find subnetworks within a randomly weighted ResNet-50 \cite{resnet}, Wide ResNet-50 \cite{wideresnet}, and ResNet-101. Notably, a randomly weighted Wide ResNet-50 contains a subnetwork which is smaller than, but matches the performance of ResNet-34. Note that for the non-dense models, \# of Parameters denotes the size of the subnetwork.}
\label{fig:params}
\end{figure}

\subsection{ImageNet \cite{imagenet} Experiments}\label{sec:imagenet}

On ImageNet we observe similar trends to CIFAR-10. As ImageNet is a much harder dataset, computationally feasible models are not overparameterized to the same degree. As a consequence, the performance of a randomly weighted subnetwork does not match the full model with learned weights.
However, we still witness a very encouraging trend -- the performance increases with the width and depth of the network. 

As illustrated by Figure \ref{fig:params}, a randomly weighted Wide ResNet-50 contains a subnetwork that is smaller than, but matches the accuracy of ResNet-34 when trained on ImageNet \cite{imagenet}. As strongly suggested by our trends, better and larger ``parent" networks would result in even stronger performance on ImageNet \cite{imagenet}. A table which reports the numbers in Figure \ref{fig:params} may be found in Section \ref{sec:imtable} of the appendix.

Figure \ref{fig:rn50} illustrates the effect of $k$, which follows an almost identical trend: $k \in [30,70]$ performs best though 30 now provides the best performance. Figure \ref{fig:rn50} also demonstrates that we significantly outperform Zhou \etal at scale (in their original work they do not consider ImageNet). For Zhou \etal on ImageNet we report the best top-1 accuracy as we find their performance degrades towards the end of training. This is the only case where we do not report last epoch accuracy.

\begin{figure}[t]
\begin{center}
\includegraphics[width=0.85\columnwidth]{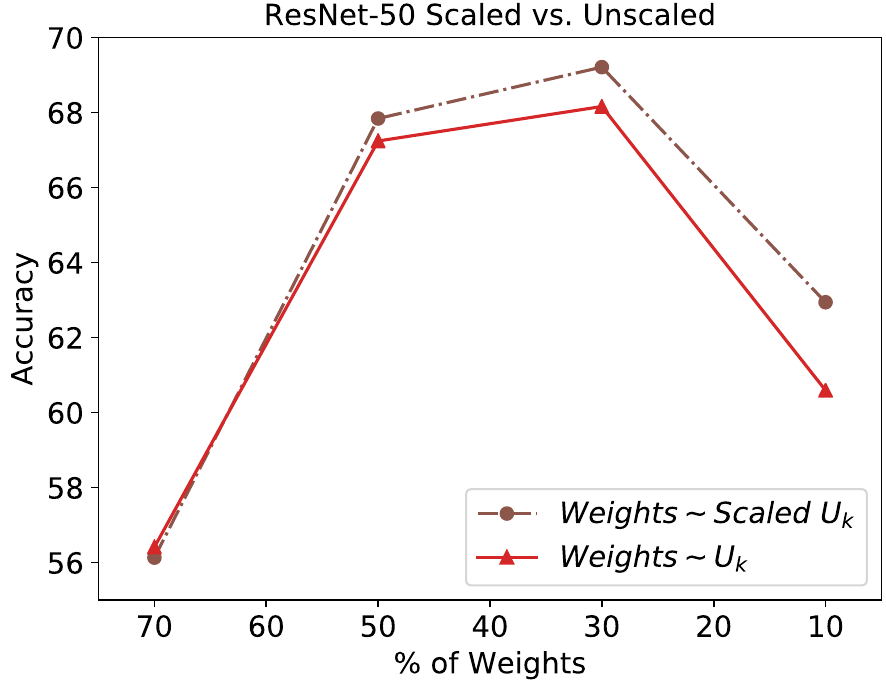}
 
\end{center}
  \caption{Examining the effect of using the ``Scaled" initialization detailed in Section \ref{sec:init} on ImageNet.}
\label{fig:rn50_scaling}
\end{figure}

The choice of the random distribution matters more for ImageNet. The ``Scaled" distribution we discuss in Section \ref{sec:init} did not show any discernable difference on CIFAR-10. However, Figure \ref{fig:rn50_scaling} illustrates that on ImageNet it is much better. Recall that the ``Scaled"
 distribution adds a factor of $\sqrt{1/k}$, which has less of an effect when $k$ approaches $100 \% = 1$. This result highlights the possibility of finding better distributions which work better for this task.

\begin{figure}[t]
\begin{center}
\includegraphics[width=0.85\columnwidth]{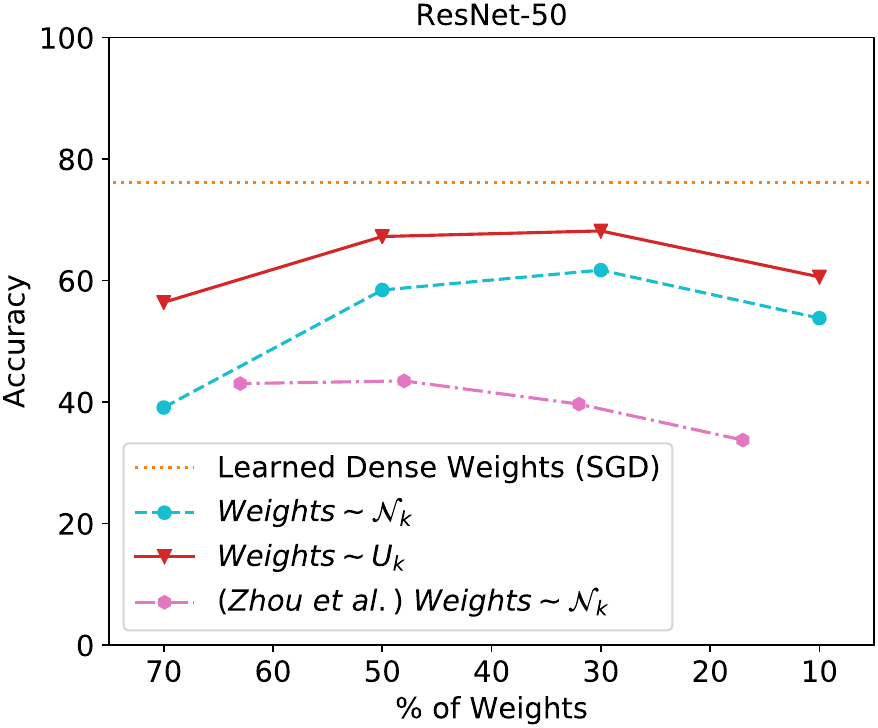}
\end{center}
  \caption{Examining the effect of $\%$ \textit{weights} on ImageNet for \alg and the method of Zhou \etal.}
\label{fig:rn50}
\end{figure}

\vspace{-1em}

\section{Conclusion}

Hidden within randomly weighted neural networks we find subnetworks with compelling accuracy. This work provides an avenue for many areas of exploration. %For example, we anticipate the development of faster algorithms, or the alternating optimization of the structure and the weights. 
Finally, we hope that our findings serve as a useful step in the pursuit of understanding the optimization and initialization of neural networks.
\small{
\subsubsection*{Acknowledgments}
We thank Jesse Dodge and Nicholas Lourie and Gabriel Ilharco for helpful discussions and Sarah Pratt for ${n \choose 2}$. This work is in part supported by NSF IIS 1652052, IIS 17303166, DARPA N66001-19-2-4031,  67102239, gifts from  Allen Institute for Artificial Intelligence, and the AI2 Fellowship for AI.
}

{\small
\bibliographystyle{ieee_fullname}
\bibliography{egbib}
}

\clearpage
 
\appendix
\begin{table*}[ht!]
  \small
  \centering
  \begin{tabular}{lllccl}
    \toprule
    Method & Model & Initialization & \% of Weights & \# of Parameters & Accuracy\\
    \midrule
    & ResNet-34 \cite{resnet} & - & - & 21.8M & 73.3\% \\
    Learned Dense Weights (SGD) & ResNet-50 \cite{resnet}& - & - & 25M & 76.1\%\\
    & Wide ResNet-50 \cite{wideresnet} & - & - & 69M & 78.1\%\\
    \midrule
    & ResNet-50 & Kaiming Normal & 30\% & 7.6M & 61.71\%\\
    \alg & ResNet-101 & Kaiming Normal & 30\% & 13M & 66.15\%\\
    & Wide ResNet-50 & Kaiming Normal & 30\% & 20.6M & 67.95\%\\
    \midrule
    & ResNet-50 & Signed Kaiming Constant & 30\% & 7.6M & 68.6\%\\
    \alg & ResNet-101 & Signed Kaiming Constant & 30\% & 13M & 72.3\%\\
    & Wide ResNet-50 & Signed Kaiming Constant & 30\% & 20.6M & 73.3\%\\
    \bottomrule
        \\
    \end{tabular}
    \caption{ImageNet \cite{imagenet} classification results corresponding to Figure \ref{fig:params}. Note that for the non-dense models, \# of Parameters denotes the size of the subnetwork.}
\label{tab:largescale}
\end{table*} 

\section{Table of ImageNet Results} \label{sec:imtable}

In Table \ref{tab:largescale} we provide a table of the results for image classification with ImageNet \cite{imagenet}. These results correspond exactly to Figure \ref{fig:params}.

\section{Additional Technical Details} \label{sec:technical-details}

In this section we first prove a more general case of Theorem 1 then provide an extension of \alg for convolutions along with code in PyTorch \cite{pytorch}, found in Algorithm \ref{alg:code}.

\subsection{A More General Case of Theorem 1} \label{sec:general-case}

\noindent \textbf{Theorem 1 (more general):} When a nonzero number of edges are swapped in one layer and the rest of the network remains fixed then the loss decreases for the mini-batch (provided the loss is sufficiently smooth).

\noindent \textit{Proof.} As before, we let $\tilde s_{uv}$ denote the score of weight $w_{uv}$ after the gradient update. Additionally, let $\tilde \I_v$ denote the input to node $v$ after the gradient update whereas $\I_v$ is the input to node $v$ before the update. Finally, let $i_1,...,i_n$ denote the $n$ nodes in layer $\ell - 1$ and $j_1,...,j_m$ denote the $m$ notes in layer $\ell$. Our goal is to show that 
\begin{align} \label{eq:goal}
    \loss\left(\tilde \I_{j_1},...,\tilde \I_{j_m} \right) < \loss\Big(\I_{j_1},..., \I_{j_m} \Big)
\end{align}
where the loss is written as a function of layer $\ell$'s input for brevity.
If the loss is smooth and $\tilde \I_{j_k}$ is close to $\I_{j_k}$ we may ignore second-order terms in a Taylor expansion:
\begin{align}
    &\loss\left(\tilde \I_{j_1},...,\tilde \I_{j_m} \right) \\
    &= \loss\left(\I_{j_1} + \left(\tilde \I_{j_1}  - \I_{j_1}\right),..., \I_{j_m} + \left(\tilde \I_{j_m}  - \I_{j_m}\right)  \right) \\
    &= \loss\left(\I_{j_1},..., \I_{j_m} \right)  + \sum_{k=1}^m \frac{\partial \loss}{\partial \I_{j_k}} \left(\tilde \I_{j_k}  - \I_{j_k}\right)
\end{align}

And so, in order to show Equation \ref{eq:goal} it suffices to show that 
\begin{align} \label{eq:goal2}
    \sum_{k=1}^m \frac{\partial \loss}{\partial \I_{j_k}} \left(\tilde \I_{j_k}  - \I_{j_k}\right) < 0.
\end{align}

It is helpful to rewrite the sum to be over edges. Specifically, we will consider the sets $\E_\textrm{old}$ and $\E_\textrm{new}$ where $\E_\textrm{new}$ contains all edges that entered the network after the gradient update and $\E_\textrm{old}$ consists of edges which were previously in the subnetwork, but have now exited. As the total number of edges is conserved we know that $|\E_\textrm{new}| = |\E_\textrm{old}|$, and by assumption $|\E_\textrm{new}| >0$.

Using the definition of $\I_k$ and $\tilde \I_k$ from Equation \ref{eq:our-update} we may rewrite Equation \ref{eq:goal2} as
\begin{align} \label{eq:goal3}
    \sum_{(i_a, j_b) \in \E_\textrm{new} } \frac{\partial \loss}{\partial \I_{j_b}}w_{i_aj_b}Z_{i_a} - \sum_{(i_c, j_d) \in \E_\textrm{old} } \frac{\partial \loss}{\partial \I_{j_d}}w_{i_cj_d}Z_{i_c}  < 0
\end{align}
which, by Equation \ref{eq:update-rule} and factoring out $1/\alpha$ becomes
\begin{align} \label{eq:goal4}
    \sum_{(i_a, j_b) \in \E_\textrm{new} } (s_{i_aj_b} - \tilde s_{i_aj_b}) - \sum_{(i_c, j_d) \in \E_\textrm{old} }  (s_{i_cj_d} - \tilde s_{i_cj_d})  < 0.
\end{align}
We now show that 
\begin{align}\label{eq:geq0}
    (s_{i_aj_b} - \tilde s_{i_aj_b}) - (s_{i_cj_d} - \tilde s_{i_cj_d}) < 0
\end{align}
for any pair of edges $(i_a, j_b) \in \E_\textrm{new}$ and $(i_c, j_d) \in \E_\textrm{old}$. Since $|\E_\textrm{new}| = |\E_\textrm{old}| > 0$ we are then able to conclude that Equation \ref{eq:goal4} holds.

As $(i_a, j_b)$ was not in the edge set before the gradient update, but $(i_c, j_d)$ was, we can conclude 
\begin{align}\label{eq:geq1}
s_{i_aj_b} - s_{i_cj_d} < 0.
\end{align}
Likewise, since $(i_a, j_b)$ is in the edge set after the gradient update, but $(i_c, j_d)$ isn't, we can conclude
\begin{align}\label{eq:geq2}
\tilde s_{i_cj_d} - \tilde s_{i_aj_b} < 0.
\end{align}

By adding Equation~\ref{eq:geq2} and Equation~\ref{eq:geq1} we find that Equation~\ref{eq:geq0} is satisfied as needed.

\subsection{Extension to Convolutional Neural Networks} \label{sec:convolutional-case}

In order to show that our method extends to convolutional layers we recall that convolutions may be written in a form that resembles Equation~\ref{eq:linear}. Let $\kappa$ be the kernel size which we assume is odd for simplicity, then for $w \in \{1,...,W\}$ and $h \in \{1,...,H\}$ we have
\begin{equation} \label{eq:conv}
\begin{split}
    \I_{v}^{w,h} = \sum_{u \in \V^{(\ell - 1)}}& \sum_{\kappa_1=1}^{\kappa} \sum_{\kappa_2 = 1}^{\kappa} \\
    &w_{uv}^{(\kappa_1,\kappa_2)} \Z_u^{\left(w+\kappa_1 -
    \left\lceil\frac{\kappa}{2}\right\rceil, h+\kappa_2 - \left\lceil\frac{\kappa}{2}\right\rceil\right)}
\end{split}
\end{equation}
where instead of ``neurons", we now have ``channels". The input $\I_v$ and output $\Z_v$ are now two dimensional and so $\Z_v^{(w,h)}$ is a scalar. As before, $\Z_v = \sigma\left(\I_v\right)$ where $\sigma$ is a nonlinear function. However, in the convolutional case $\sigma$ is often batch norm \cite{batchnorm} followed by ReLU (and then implicitly followed by zero padding).

%##################################################################################################

%##################################################################################################

Instead of simply having weights $w_{uv}$ we now have weights $w_{uv}^{(\kappa_1,\kappa_2)}$ for $\kappa_1 \in \{1,...,\kappa\}$, $\kappa_2 \in \{1,...,\kappa\}$. Likewise, in our \alg Algorithm we now consider scores $s_{uv}^{(\kappa_1,\kappa_2)}$ and again use the top $k$\% in the forwards pass. As before, let $h\left(s_{uv}^{(\kappa_1,\kappa_2)}\right) = 1$ if $s_{uv}^{(\kappa_1,\kappa_2)}$ is among the top $k\%$ highest scores in the layer and $h\left(s_{uv}^{(\kappa_1,\kappa_2)}\right) = 0$ otherwise. Then in \alg we are performing a convolution as
\begin{align} \label{eq:conv2}
\begin{split}
    \I_{v}^{w,h} = &\sum_{u \in \V^{(\ell - 1)}} \sum_{\kappa_1=1}^{\kappa} \sum_{\kappa_2 = 1}^{\kappa} \\
    &w_{uv}^{(\kappa_1,\kappa_2)} \Z_u^{\left(w+\kappa_1 -
    \left\lceil\frac{\kappa}{2}\right\rceil, h+\kappa_2 - \left\lceil\frac{\kappa}{2}\right\rceil\right)} h\left(s_{uv}^{(\kappa_1, \kappa_2)}\right)
\end{split}
\end{align}
which mirrors the formulation of \alg in Equation~\ref{eq:our-update2}.  In fact, when $\kappa = W = H = 1$ (\ie a 1x1 convolution on a 1x1 feature map) then Equation \ref{eq:conv2} and Equation \ref{eq:our-update2} are equivalent. 

The update for the scores is quite similar, though we must now sum over all spatial (\ie $w$ and $h$) locations as given below:
\begin{equation} \label{eq:update-rule3}
\begin{split}
    & s_{uv}^{(\kappa_1, \kappa_2)} \gets s_{uv}^{(\kappa_1, \kappa_2)} \\
    & \ \ \ \ \ - \alpha \sum_{w=1}^W \sum_{h=1}^H \pdv{\loss}{\I_{v}^{w,h}} w_{uv}^{(\kappa_1, \kappa_2)} \Z_u^{\left(w+\kappa_1 -
    \left\lceil\frac{\kappa}{2}\right\rceil, h+\kappa_2 - \left\lceil\frac{\kappa}{2}\right\rceil\right)}
\end{split}
\end{equation}

In summary, we now have $\kappa^2$ edges between each $u$ and $v$.
The PyTorch \cite{pytorch} code is given by Algorithm~\ref{alg:code}, where $h$ is \texttt{GetSubnet}. The gradient goes straight through $h$ in the backward pass, and PyTorch handles the implementation of these equations.

\begin{algorithm}[t]
\caption{PyTorch code for an \alg Conv.}
\label{alg:code}
\definecolor{codeblue}{rgb}{0.25,0.5,0.5}
\definecolor{codeblue2}{rgb}{0,0,1}
\lstset{
  backgroundcolor=\color{white},
  basicstyle=\fontsize{7.2pt}{7.2pt}\ttfamily\selectfont,
  columns=fullflexible,
  breaklines=true,
  captionpos=b,
  commentstyle=\fontsize{7.2pt}{7.2pt}\color{codeblue},
  keywordstyle=\fontsize{7.2pt}{7.2pt}\color{codeblue2},
%  frame=tb,
}
\begin{lstlisting}[language=python]
class GetSubnet(autograd.Function):
    @staticmethod
    def forward(ctx, scores, k):
         # Get the subnetwork by sorting the scores and using the top k%
        out = scores.clone()
        _, idx = scores.flatten().sort()
        j = int((1-k) * scores.numel())

        # flat_out and out access the same memory.
        flat_out = out.flatten()
        flat_out[idx[:j]] = 0
        flat_out[idx[j:]] = 1

        return out

    @staticmethod
    def backward(ctx, g):
        # send the gradient g straight-through on the backward pass.
        return g, None
        

class SubnetConv(nn.Conv2d):
    # self.k is the % of weights remaining, a real number in [0,1]
    # self.popup_scores is a Parameter which has the same shape as self.weight
    # Gradients to self.weight, self.bias have been turned off.
    def forward(self, x):
        # Get the subnetwork by sorting the scores.
        adj = GetSubnet.apply(
                    self.popup_scores.abs(), self.k)
        # Use only the subnetwork in the forward pass.
        w = self.weight * adj
        x = F.conv2d(
            x, w, self.bias, self.stride, self.padding, self.dilation, self.groups
        )
        return x
        
\end{lstlisting}
\end{algorithm}

\begin{table*}[ht!]
  \small
  \centering
  \begin{tabular}{cllcccl}
    \toprule
    Subnetwork Type & Model & Initialization & \% of Weights & \# of Parameters & Top 1 Accuracy & Top 5 Accuracy\\
    \midrule
            & ResNet-50 \cite{resnet}& Kaiming Normal & 30\% & 7.6M & 73.6\% & 91.6\%\\
    Learned & ResNet-50 \cite{resnet}& Signed Constant & 30\% & 7.6M & 73.7\% & 91.5\%\\
            & Wide ResNet-50 \cite{wideresnet}& Kaiming Normal & 30\% & 20.6M & 76.8\% & 93.2\%\\
            & Wide ResNet-50 \cite{wideresnet}& Signed Constant & 30\% & 20.6M & 76.9\% & 93.3\%\\
            \midrule
            & ResNet-50 \cite{resnet}& Kaiming Normal & 30\% & 7.6M & 73.5\% & 91.5\%\\
    Random  & Wide ResNet-50 \cite{wideresnet} & Kaiming Normal & 30\% & 20.6M & 76.5\% & 93.1\%\\
            & ResNet-101 \cite{resnet} & Kaiming Normal & 30\% & 13M & 76.1\% & 93.0\%\\
    \bottomrule
        \\
    \end{tabular}
    \caption{ImageNet \cite{imagenet} classification results after training the discovered subnetworks. Surprisingly, the
    accuracy is not substantially better than training a random subnetwork. This suggests that the good performance of these subnetworks does not explain the lottery phenomena described in \cite{lth}.}
\label{tab:lottery}
\end{table*}

\section{Additional Experiments}

\subsection{Resnet18 on CIFAR-10}

\begin{figure}[]
    \centering
    \includegraphics[width=\columnwidth]{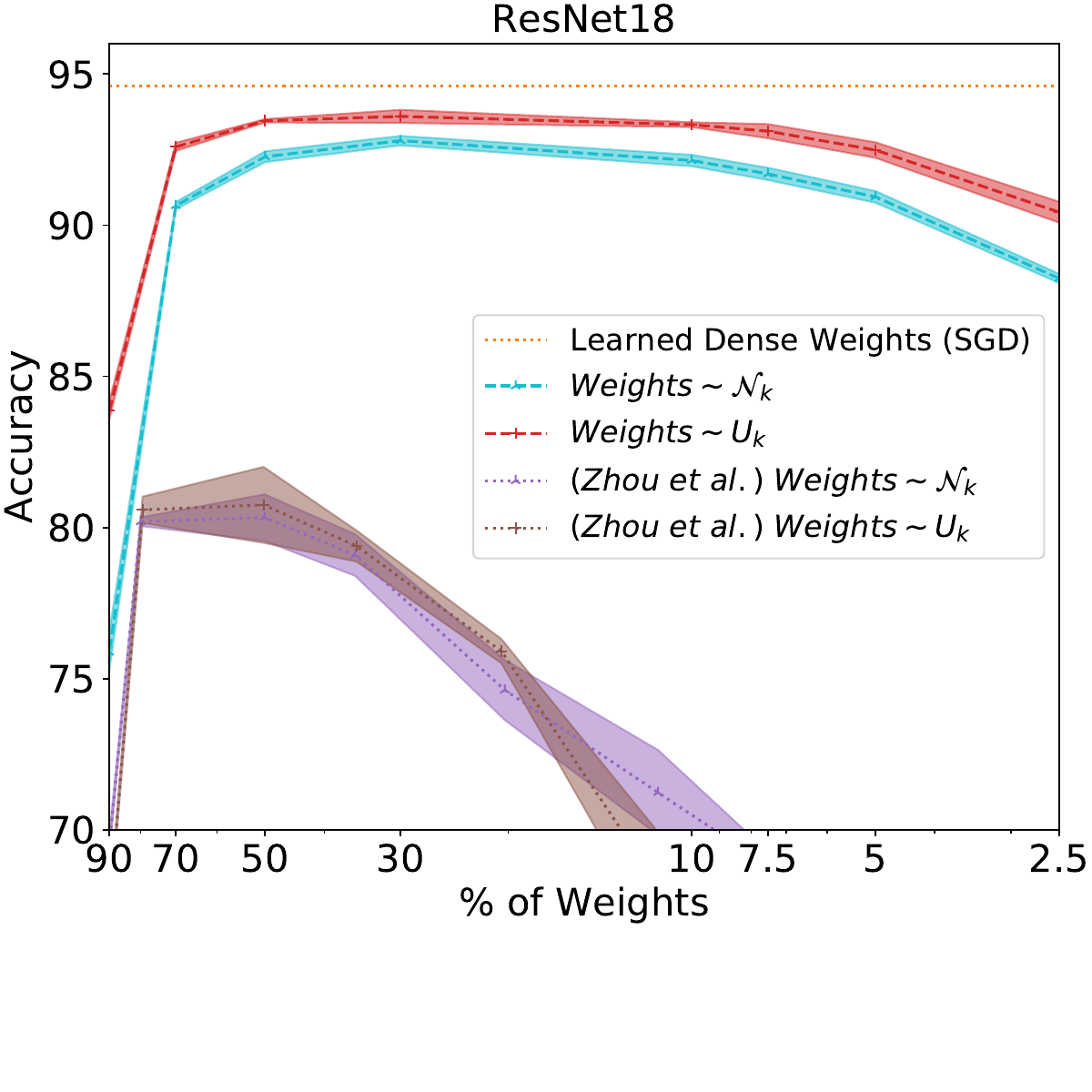}
    \caption{Repeating the experiments from Figure \ref{fig:conv2468} with ResNet18 on CIFAR.}
    \label{fig:resnet18}
\end{figure}

In figure \ref{fig:resnet18} we experiment with a more advanced network architecture on CIFAR-10. 

\subsection{Are these subnetworks lottery tickets?}

What happens when we train the weights of the subnetworks form Figure \ref{fig:params} and Table \ref{tab:largescale} on ImageNet? They do not train to the same accuracy as a dense network, and do not perform substantially better than training a random subnetwork. This suggests that the good performance of these subnetworks at initialization does not explain the lottery phenomena described in \cite{lth}. The results can be found in Table \ref{tab:lottery}, where we again use the hyperparameters found on NVIDIA's public github example repository for training ResNet \cite{repo}.

\end{document}